\documentclass[a4paper, 10pt, conference]{ieeeconf}      % Use this line for a4
\usepackage{FG2019}
\usepackage{multirow}
\IEEEoverridecommandlockouts                              % This command is only
\usepackage{graphics} % for pdf, bitmapped graphics files
\usepackage{graphicx}
\usepackage{subcaption}
\usepackage[noadjust]{cite}
\def\FGPaperID{150} % *** Enter the FG 2019 Paper ID here
\title{\LARGE \bf
Revisiting a single-stage method for face detection}

\author{\parbox{16cm}{\centering
    {\large Nguyen Van Quang$^{1,2}$, Hiromasa Fujihara$^{2}$}\\
    {\normalsize
    $^1$ Faculty of Information Sciences, Tohoku University, Sendai, Japan\\
    $^2$ Laboro.AI Inc., Tokyo, Japan}}
    \thanks{}% <-this % stops a space
}

\begin{document}
\ifFGfinal
\thispagestyle{empty}
\pagestyle{empty}
\else
\author{Anonymous FG 2019 submission\\ Paper ID, 150 \\}
\pagestyle{plain}
\fi
\maketitle
%%%%%%%%%%%%%%%%%%%%%%%%%%%%%%%%%%%%%%%%%%%%%%%%%%%%%%%%%%%%%%%%%%%%%%%%%%%%%%%%
\begin{abstract}
Although accurate, two-stage face detectors usually require more inference time than single-stage detectors do. This paper proposes a simple yet effective single-stage model for real-time face detection with a prominently high accuracy. We build our single-stage model on the top of the ResNet-101 backbone and analyze some problems with the baseline single-stage detector in order to design several strategies for reducing the false positive rate. The design leverages the context information from the deeper layers in order to increase recall rate while maintaining a low false positive rate. In addition, we reduce the detection time by an improved inference procedure for decoding outputs faster. The inference time of a VGA ($640{\times}480$) image was only approximately 26 ms with a Titan X GPU. The effectiveness of our proposed method was evaluated on several face detection benchmarks (Wider Face, AFW, Pascal Face, and FDDB). The experiments show that our method achieved competitive results on these popular datasets with a faster runtime than the current best two-stage practices.
\end{abstract}
%%%%%%%%%%%%%%%%%%%%%%%%%%%%%%%%%%%%%%%%%%%%%%%%%%%%%%%%%%%%%%%%%%%%%%%%%%%%%%%%

\section{INTRODUCTION}
\baselineskip 0.9\normalbaselineskip
Face detection has attracted much attention as it constitutes the fundamental step of many common face-related tasks. Since the pioneering work of Viola-Jones \cite{viola}, face detection has progressively improved \cite{cascade2010,shapemodel, dpm3}, but still relying on laborious feature engineering to train the face detectors. However, in challenging datasets such as the recently introduced Wider Face \cite{widerface}, these approaches show the non-robustness to a wide range of facial variations. Recently, the CNN-based methods have been increasingly adopted to deal with the facial variations and have surpassed the former methods. Nevertheless, there is a trade-off between computational cost and accuracy in face detection. For instance, the HR detector \cite{hr} needs to forward an image pyramid into a shared CNN model which consumes more than 1 second for accurate results.

Inherited from object detection, two main approaches have been applied successfully to face detection, namely, one-stage (single-stage) and two-stage methods. Two-stage methods follow a common pipeline: (1) produce a set of region proposals with their local features (or pixels), (2) pass them to the second network for classifying and regressing the bounding boxes of the detected faces. Though very accurate, these systems require numerous intensive computations. For example, Chenchen Zhu et al. \cite{ssf} recently proposed a two-stage detector which obtains the state-of-the-art performance on face benchmarks but requires about 150ms to proceed an image of size $600{\times}1000$. On the other hand, single-stage methods extract the feature maps of convolutional layers at several depths from a base network, in which each layer is associated with a set of predefined anchors. Convolutions are then performed on these feature maps for classification and regression tasks directly. Although these methods significantly improve the runtime of object detection over two-stage methods while delivering comparative performance, applying single-shot detectors (SSD) \cite{ssd} directly to face detection does not yield the acceptable performance \cite{ssf}. However,  SFD \cite{sfd} and SSH \cite{ssh}, the one-stage models proposed for face detection using VGG-16 \cite{vgg} as a backbone framework with several improvements, could obtain very competitive results, outperforming even the top ResNet-based models. 

Inspired by SSD, SFD and SSH, we build a simple single-stage model with the ResNet-101 backbone for face detection as a baseline. After investigating the baseline model, we attribute the high rate of false positives to the following causes.

\textbf{Lack of context information.} First, small faces appear to lack deterministic facial parts due to their low resolution. Moreover, the corresponding feature maps of small faces have less context information in the shallower anchor-associated layers than in the deeper layers. Consequently, small-scale faces are difficult to detect. According to \cite{hr}, large receptive fields can increase the context information for detecting small faces. 

\textbf{Very large receptive field.} In contrast to the above, excessively large receptive fields might provide redundant information for detection which increases the false positive rate. For a given stride, the the anchor-associated layers in ResNet-101 based networks have larger receptive fields than VGG-16 based networks do. In SFD, the anchor sizes and layouts are designed reasonably such that the anchor area matches the receptive field for each stride.

\textbf{Shared feature maps for the classification and regression tasks in detection.} In SSD, two conv layers are performed directly on the extracted feature maps to do both classification and regression tasks. Therefore, the network might not easily and immediately learn the mappings and balance the losses in both tasks. 

\textbf{Dense anchors for each ground truth box, and for each cell in the feature maps.} The number of anchors assigned to small faces was increased by the matching strategy proposed elsewhere \cite{sfd,ssf}. Moreover, there are several anchors of different aspect ratios per cell in SSD. Although these strategies help increase the recall rate, it probably also contributes the high false positive rate. 

All of the above factors lessen the capacity of the ResNet backbone in single-stage models. Herein, we propose several simple but efficient strategies that reduce the false positive rate of our single-stage model with a ResNet-101 base for face detection. Overall, reducing the false positives, especially those with high confidence scores, boosts the average precision (AP) score of the model. After implementing these strategies, our method achieved the \textbf{state-of-the-art results} on AFW, PASCAL face, FDDB, and also competitive results on WIDER FACE at faster runtime speed than the current best two-stage methods. We also improve the decoding procedure to make the inference faster up to about 10\% with the speed of 38 fps for VGA images of size $640{\times}480$.
\baselineskip 1.0\normalbaselineskip

\baselineskip 0.92\normalbaselineskip
%%%%%%%%%%%%%%%%%%%%%%%%%%%%%%%%%%%%%%%%%%%%%%%%%%%%%%%%%%%%%%%
\section{RELATED WORK}
\subsection{Face detection}
In the past decades, Viola and Jones \cite{viola} pioneered a real-time face detector, using Haar-like features to train a cascade of Adaboost models. After that, face detection has continuously improved in various works from \cite{cascade2010, shapemodel}, to deformable part models (DPM) \cite{dpm1,dpm2, dpm3}. Most of these methods train traditional classifiers using a selective set of hand-crafted features. Recently, CNN-based methods have significantly boosted the performance of face detection over traditional methods which struggle to deal with facial variations. These recent state-of-the-art face detectors can be divided into two main approaches: one-stage detectors and two-stage detectors. We follow the first approach to design our face detector.

\subsection{Single-stage networks for face detection}
Several works follow the design of one-stage detection architecture, having achieved very remarkable results on common benchmarks for face detection. SFD \cite{sfd} adopts the VGG-16 but tiling and assigning the anchors more tightly for small faces to increase the recall score. Mahyar Najibi et al. \cite{ssh} proposed a network called SSH which removes the fully connected layers from the VGG-16 base network, and provides the context information for detection module by fusing several convolutional layers. To improve the detection of occluded faces, Wang et al. \cite{fan} proposed FAN model which employs the RetinaNet \cite{retina} using feature pyramid integrated with the attention mechanism.

%%%%%%%%%%%%%%%%%%%%%%%%%%%%%%%%%%%%%%%%%%%%%%%%%%%%%%%%%%%%%%%
\section{OUR PROPOSED NETWORK}

\subsection{The network architecture}
Following the SFD design, we also build a single-stage framework which is robust to face scales. We extract the feature maps of different convolutional layers from the network. Each cell in the feature map from each layer is associated with one square anchor (aspect ratio of 1:1). The anchor size from each extracted conv layer increases exponentially with the power of 2, ranging from 16 to 512 pixels. As implemented in SFD, the strides of anchors are also kept unchanged to guarantee that faces of different scales are assigned to the equally adequate number of matched anchors from the feature maps. The model consists of the feature extractor, fusion modules and detection modules. The feature extractor is constructed by several of following layers:

\textbf{Base convolutional layers.} We design a very deep CNN network using ResNet-101 \cite{resnet} which is well known for producing highly representative features for extraction. All the layers of ResNet-101 before the first fully connected layers are retained as our feature extractor.

\textbf{Extra convolutional layers}. For detecting large faces, we add the extra convolutional layers to the ResNet-101 base in order. The additional $Conv\_6$ and $Conv\_7$ blocks are sequential blocks of two $3{\times}3$ conv layers of $N=256$ filters, in which the latter layer with a stride of 2  is followed by ReLU activation.

\textbf{Detection convolutional layers.} For further detection, we extract feature maps of several layers from base convolutional layers and extra convolutional layers as anchor-associated layers for further detection. These extracted layers ($Conv2\_3$, $Conv3\_3$, $Conv4\_3$, $Conv5\_3$, $Conv6\_2$, and $Conv7\_2$) are the last conv layers in their corresponding blocks. We convolve the selected layers with $1{\times}1$ conv layer of 256 filters to ensure that all the extracted layers have $256$ channels if their number of channels differs from 256.

\setlength\belowcaptionskip{-3ex}
\begin{figure}[t]
    \centering
    \includegraphics[width=0.47\textwidth]{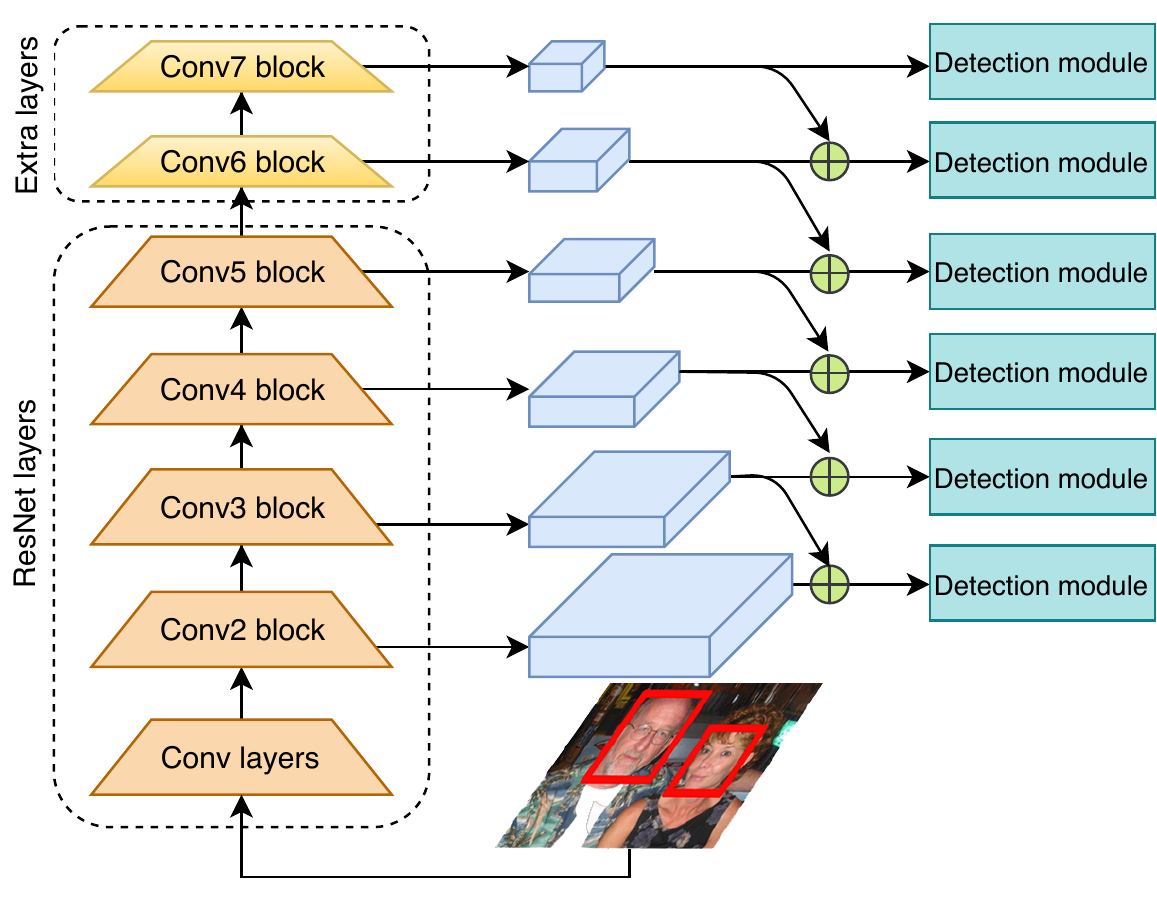}
    \caption{The proposed model architecture.}
\end{figure}

\subsection{Strategies for reducing the false positives}

\subsubsection{Fusing features from the higher layer}
Although the ResNet-101 backbone provides the larger receptive field for each detection convolutional layer than VGG-16 base, the shallower layers still lack context information for the detection. Fusing the features from higher layers is the common practice to enlarge the receptive fields and increase the context information for shallower layers. For instance, RetinaNet \cite{retina} designed the feature pyramid architecture which allows a layer to enjoy the flow of context information from all higher layers in the hierarchy. For face detection, the feature pyramid architecture might provide redundant context features for small face detection, increasing the false positive rate. In order to increase the context information of the feature maps in the lower layers, we fuse the current detection layer with only one consecutive higher detection layer in the hierarchical order. Therefore, we upsample the higher detection layer by factor of 2 and fuse it with the current detection layer by element-wise addition as depicted Figure 2(a). This strategy approximately doubles the receptive field of the corresponding layer approximately which provides sufficient context information for detecting small faces while reducing the risk of false positive detection. 

\setlength\belowcaptionskip{-0.2ex}
\begin{figure}[t]
     \centering
    \begin{subfigure}[t]{0.45\textwidth}
        \raisebox{-\height}{\includegraphics[width=\textwidth]{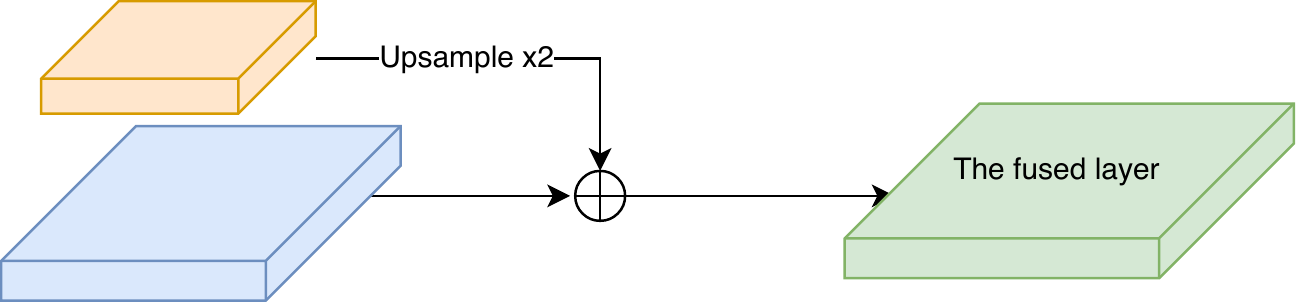}}
        \caption{Fusion module}
    \end{subfigure}
    \hfill
    \begin{subfigure}[t]{0.45\textwidth}
        \raisebox{-\height}{\includegraphics[width=\textwidth]{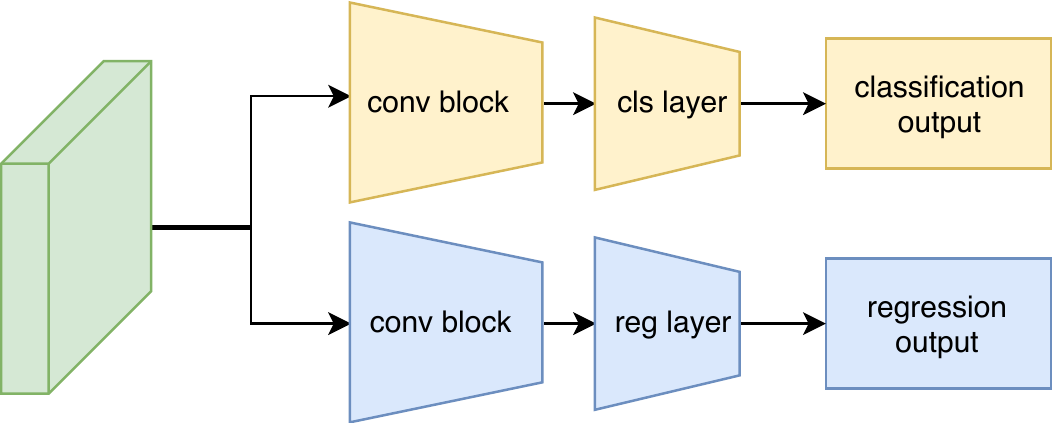}}
        \caption{Detection module}
    \end{subfigure}
    \caption{Fusion and Detection modules.}
\end{figure}

\subsubsection{Shared features for detection tasks}
In SFD, the detection layers are directly convolved with classification and regression filters. This makes the network harder to learn and optimize the loss of two tasks. To avoid this problem, we separate the detection tasks into two branches. Prior to the classification and regression layer, we add a sequential block including several $3{\times}3$ conv layers of 256 filters for each layer on each branch as implemented in RetinaNet. For each anchor, we need to regress the 4 offsets related to its coordinates and $N$ confidence scores for classification (as we have only the background label and the face label, we set $N = 2$). Therefore, we apply $3{\times}3$ convolutional layers of $4$ and $2$ filters on the regression branch and classification branch, respectively. Figure 2(b) describes the detection module.

\subsubsection{Anchor assigning strategy}
In the training phase, we need to assign a ground truth box of faces to an anchor or a set of anchors. In \cite{sfd,ssf}, they adopt a strategy for matching which anchors to a ground truth box by decreasing the threshold of Jaccard overlap between a face bounding box and anchors. This strategy significantly increases the number of anchors per a ground truth faces, especially the small faces. Although this strategy increases the recall rate for small faces, it produces more false positives. Following the same matching strategy as SFD, we increase the number of anchors per small faces but maintaining the threshold of Jaccard overlap in the first step at $0.5$. In the second step, at most 4 anchors that most overlap the hard faces (i.e., have the highest Jaccard overlap) are selected as the matched anchors. The remaining anchors are labeled as negative.

\subsection{Improved decoding strategy}
We adopt an efficient decoding strategy during the inference although single-stage models like SSD enjoy very fast detection speed. During the inference, transfering raw outputs of detection modules from the GPU to CPU also incurs a time burden. For an image with the size of $640{\times}640$, we need to transfer the output tensor with size of $37,500$ which corresponds to the number of generated anchors, in which only few of the anchors are associated with the detected faces. Hence, we compute the classification scores for all the anchors, and keep only the feature maps corresponding to anchors with confidence scores greater than a threshold $\lambda_1 = 0.1$. Then, we regress the offsets for coordinates of bounding boxes corresponding to these anchors and transfer them to the CPU for further steps. The overlapped bounding boxes are filtered by non-maximum suppression with a threshold of $\lambda_2 = 0.3$. The improved strategy decreases the inference time by approximately 10\%, improving the runtime to $36$ fps on VGA images.

%%%%%%%%%%%%%%%%%%%%%%%%%%%%%%%%%%%%%%%%%%%%%%%%%%%%%%%%%%%%%%%%%%%%%%%% 
\section{EXPERIMENTS AND RESULTS}
\subsection{Experimental setup}
\textbf{Training dataset.} The Wider Face dataset \cite{widerface} comprises 32,203 images of 393,703 annotated faces with widely varying scales, light conditions, expressions, poses and occlusions. The dataset is divided into train set, validation set, and test set (ratio 40:10:50). The validation and test sets are split into three subsets with varying detection-difficulty levels: easy, medium, and hard. We trained our model on the Wider Face train set, and evaluated on the Wider Face validation and test sets. Following \cite{sfd}, we performed color jittering and selected a random square crop with a size ratio from $0.3$ to $1.0$ (relative to the original image's shortest side). The cropped patch was resized into to $640{\times}640$, along with random horizontal flip.

\textbf{Loss function.} The multi-task loss was formulated as follows:
\begin{equation}
    L = \sum_{i}\frac{1}{N_{cls}} L_{cls}(p_i, p_i^*) + \lambda \sum_{i}\frac{1}{N_{reg}}L_{reg}(t_i, t_i^*)
\end{equation}
where $i$ is the index of an anchor and $p_i$ denotes the predicted probability that anchor $i$ is a face, $t_i$ denotes the vector of predicted offsets for anchor $i$ while $p_i^*$, $t_i^*$ are the ground truth label and offsets, respectively.  

We computed the cross entropy loss for the classification task $L_{cls}(p_i, p_i^*)$ over two classes (background and face), and divided it by the number of anchors $N_{cls}$ taken account into the loss. During the training process, the number ratio of negative examples (background) to positive examples (faces) was maintained at $3:1$ via online negative hard mining procedure. The regression loss $L_{reg}$ is a smooth $L1$ loss, and $N_{reg}$ is the number of anchors assigned to the ground truth ($p_i^* = 1$). We set $\lambda = 4$ as to optimally balance the two losses.

\textbf{Training settings.} We train our model with the ResNet module pre-trained on Imagenet dataset \cite{resnet} with the batch size of 6 on 1 GPU. SGD with momentum of $0.9$ and weight decay of $1\mathrm{e}{-4}$ is used as our optimizer. The model was trained in 300 full epochs with the initial learning rate at $1\mathrm{e}{-3}$ which was divided by $10$ when the multi-task loss $L$ plateaued. Meanwhile, the ResNet-101 based single-stage baseline was constructed with the same feature extractor as our model and the same training settings but not applied some proposed strategies. The details of the baseline will be presented in the supplementary part.

\begin{figure}[t]
    \centering
    \includegraphics[width=0.4\textwidth]{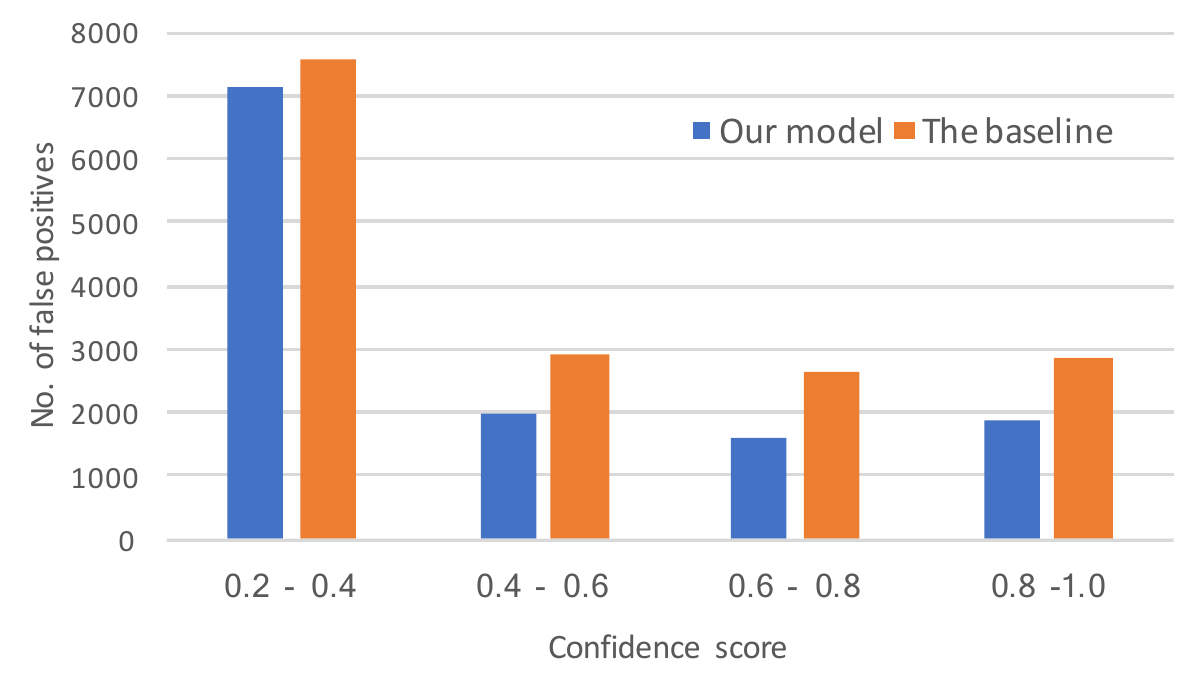}
    \caption{No. of false positives at different confidence scores of our model and the baseline on the Wider Face val set.}
\end{figure}

\begin{figure*}[t]
     \centering
    \begin{subfigure}[t]{0.3\textwidth}
        \raisebox{-\height}{\includegraphics[height=0.69\textwidth]{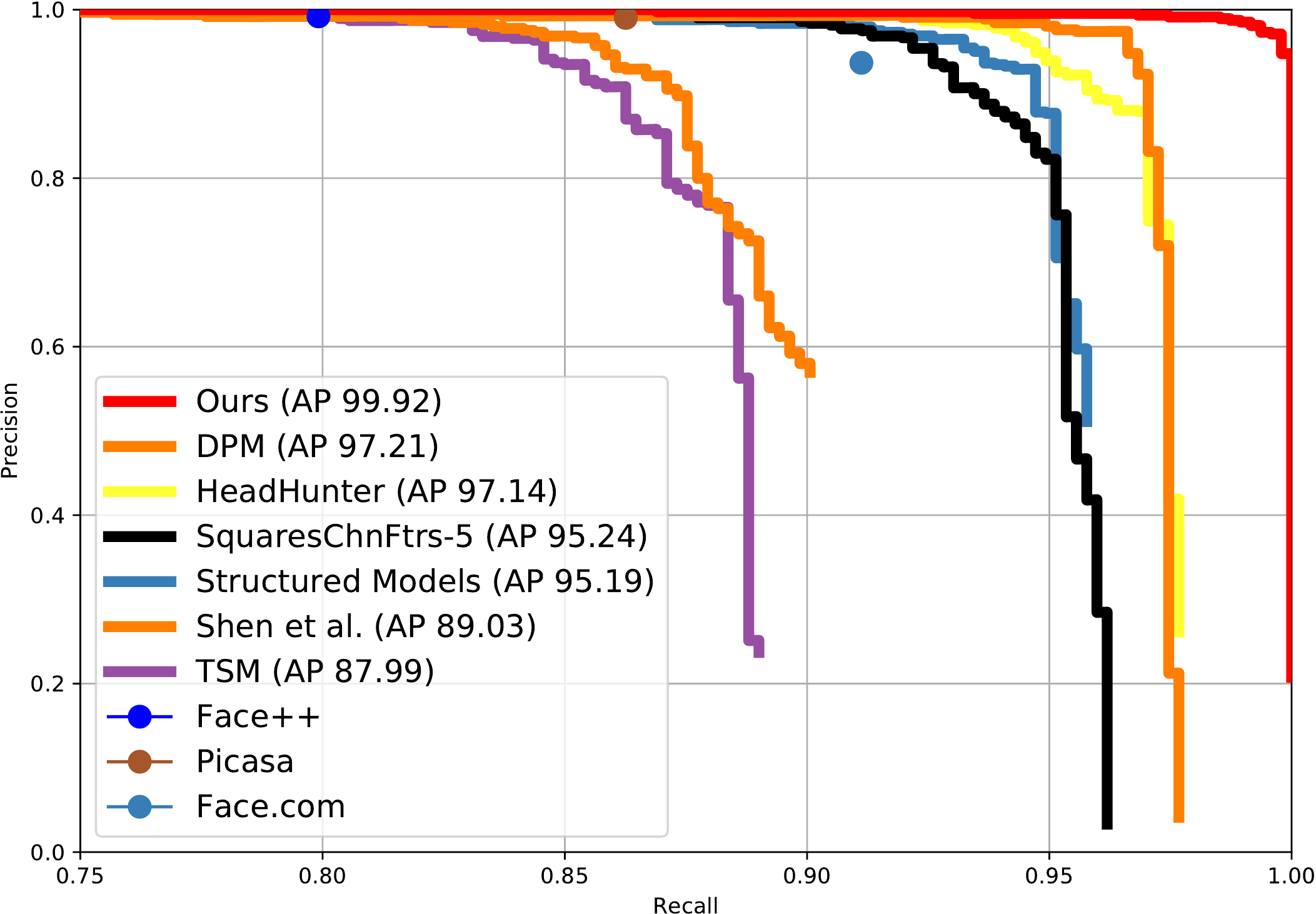}}
        \caption{Precision-recall curves on AFW}
    \end{subfigure}
    \hfill
    \begin{subfigure}[t]{0.3\textwidth}
        \raisebox{-\height}{\includegraphics[height=0.69\textwidth]{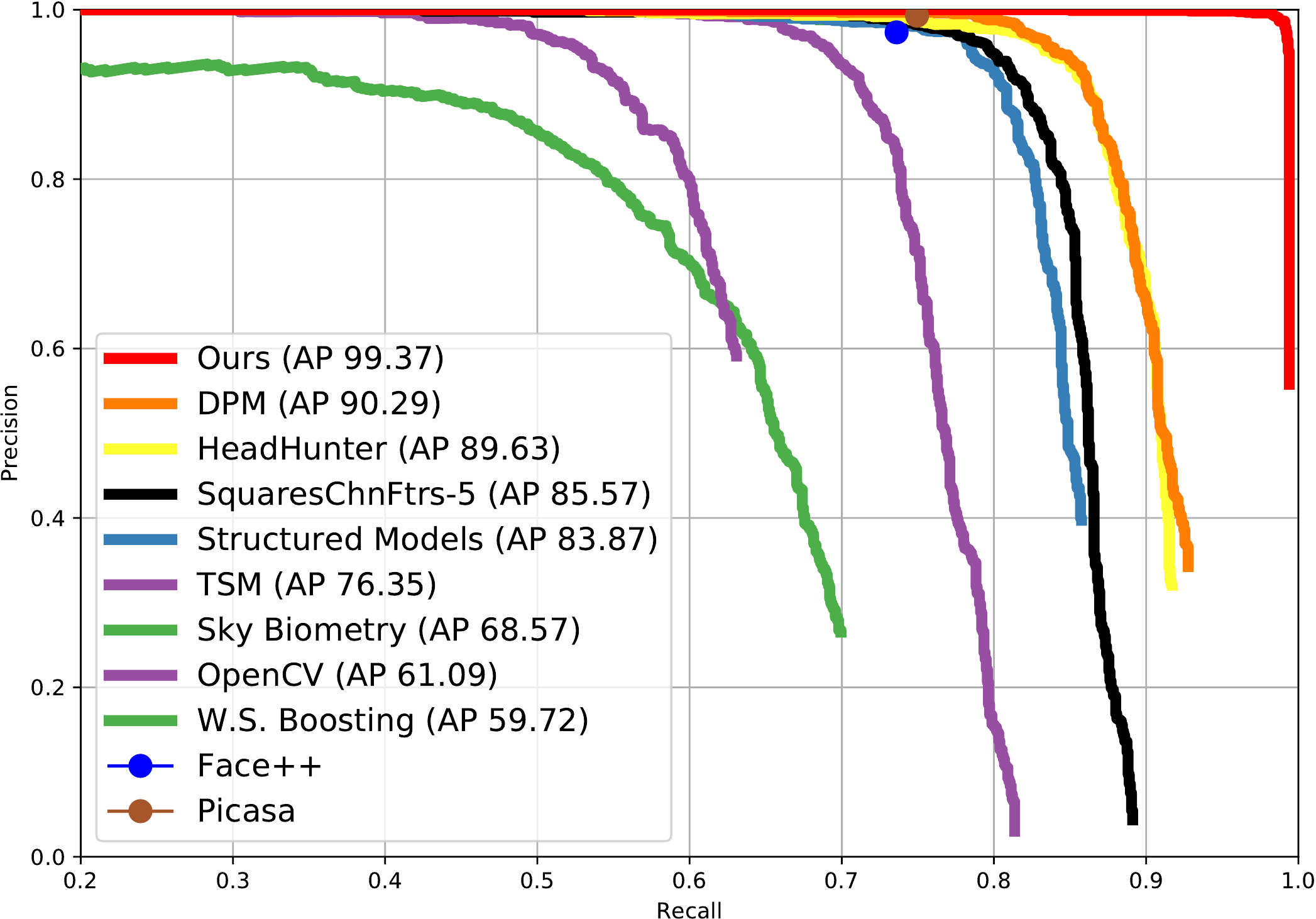}}
        \caption{Precision-recall curves on Pascal Face}
    \end{subfigure}
    \hfill
    \begin{subfigure}[t]{0.3\textwidth}
        \raisebox{-\height}{\includegraphics[height=0.69\textwidth]{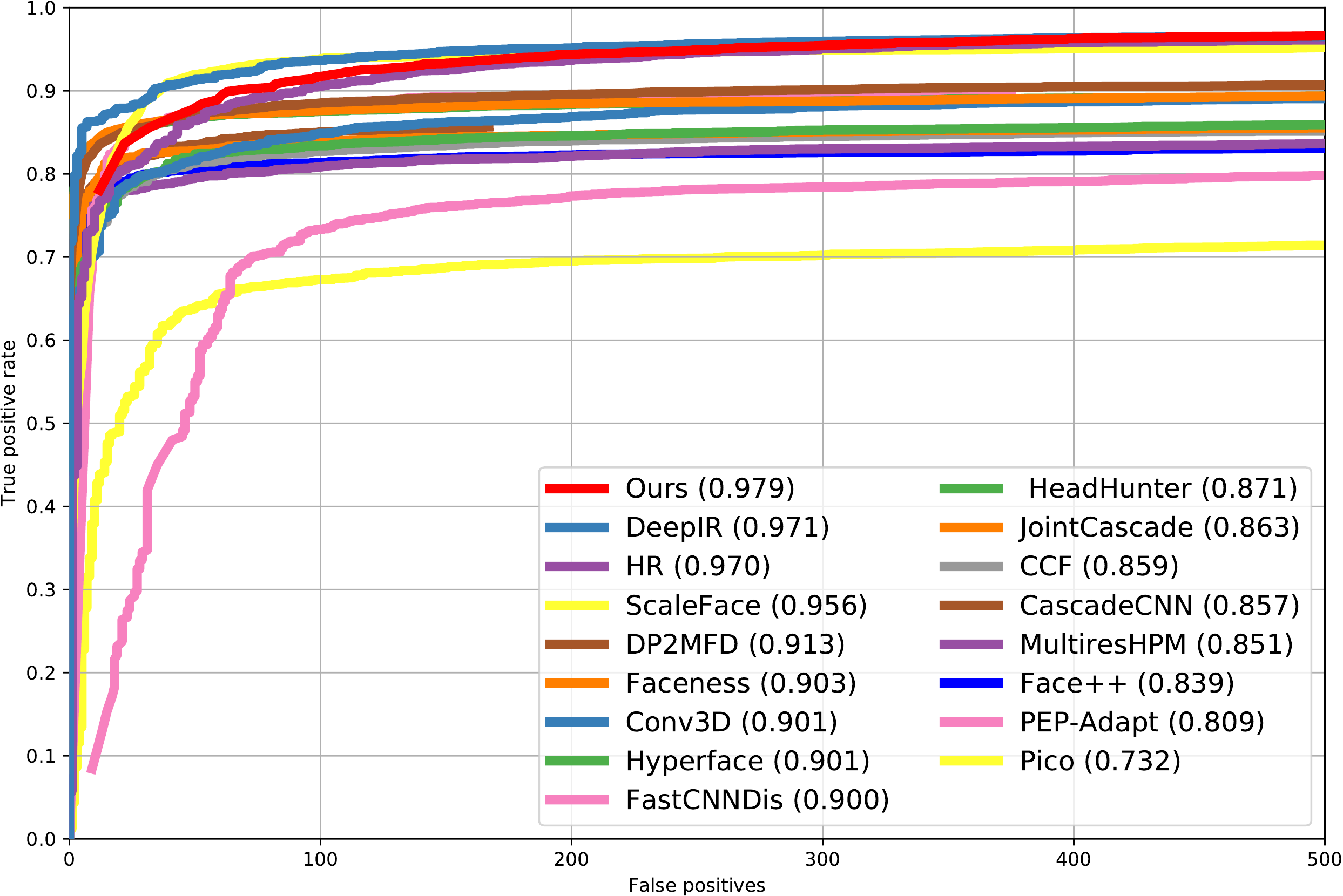}}
        \caption{Discontinuous ROC curves on FDDB}
    \end{subfigure}
    \caption{Results on common face detection benchmarks}
\end{figure*}

\subsection{Results on Common benchmarks}

\setlength\belowcaptionskip{-2ex}
\begin{table}
\caption{The AP scores of different methods on the Wider Face Test and Validation set}
\begin{center}
 \begin{tabular}{|c| c| c| c|} 
 \hline
 \multirow{2}{*}{Method} & \multicolumn{3}{c|}{AP Test (Val)\%}\\ 
 \cline{2-4}
   & Easy & Medium & Hard \\ [0.5ex] 
 \hline
 Faceness \cite{bmf6} & 71.1 (71.3) & 60.4 (63.4) & 31.5 (34.5)\\
 LDCF+ \cite{bmf4} & 79.7 (79.0) & 77.2 (76.9) & 56.4 (52.2) \\
 MT-CNN \cite{bmf3}& 85.1 (84.8) & 82.0 (82.5) & 60.7 (59.8)\\ 
 CMS-RCNN \cite{bmf50} & 90.2 (89.9) & 87.4 (87.4) & 64.3 (62.4)\\
 ScaleFace \cite{bmf2} & 86.7 (86.8) & 86.6 (86.7) & 76.2 (77.2)\\
 HR \cite{hr} & 92.3 (92.5) & 91.0 (91.0) & 81.9 (80.6) \\
 SFD \cite{sfd} & 93.5 (93.7) & 92.1 (92.5) & 85.8 (85.9) \\
 SSH \cite{ssh} & 92.7 (93.1) & 91.5 (92.1) & 84.4 (84.5)\\ 
 Detector in \cite{ssf} & 94.9 (94.9) & 93.5 (93.3) & 86.5 (86.1)\\
 \hline
 \textbf{Ours} & \textbf{94.0 (94.2)} & \textbf{92.8 (92.8)} & \textbf{85.9 (86.1)} \\
 \hline
\end{tabular}
\end{center}
\end{table}
\setlength{\textfloatsep}{0.1cm}
\textbf{Wider Face} To demonstrate the effectiveness of our proposal, Figure 3 compares the false positves of our proposed model and the ResNet baseline. The detection result was divided into three subsets with different image size thresholds. Our model is evaluated on the Wider Face validation and test sets following the standard evaluation protocol.  Table I compares the average precision scores (AP) obtained by our model and several face detectors in \cite{ssf}, SFD \cite{sfd}, SSH \cite{ssh}, ScaleFace \cite{bmf2}, HR \cite{hr}, CMS-RCNN \cite{bmf50}, Multitask Cascade CNN \cite{bmf3}, LDCF+ \cite{bmf4}, and Faceness \cite{bmf6} on Wider Face test and validation sets. Among the compared methods, our detector achieved the second best AP scores on all face cases.

To show the consistency of our model's performance, we evaluated the model on other common dataset benchmarks FDDB \cite{fddb}, AFW \cite{afw}, and Pascal Faces \cite{pascalface}. Following \cite{ssh}, we resize one side of images to 640 and keep the image aspect ratio unchanged. Interestingly, our model achieved state-of-the-art results on small resized images of these datasets without constructing an image pyramid.

\textbf{AFW} The dataset comprises 205 images in which 473 faces are annotated. We compared our model with several popular methods \cite{dpm1, bm2, dpm3, bm4} and commercial face detectors (Face.com, Picasa and Face++). Our model achieved a state-of-the-art result with an AP score of 99.92\%. Figure 4(a) depicts the comparisons.

\textbf{Pascal Face} This dataset comprises 851 images in which 1,335 faces are annotated. We compared our model with some popular methods \cite{dpm1,bm2,dpm3} and commercial face detectors (Picasa, Face++). Again, our model outperformed the other detectors by a large margin with an AP score of 99.37\% (Figure 4(b)). 

\textbf{FDDB} This dataset comprises of 2,845 images in which 5,171 faces are annotated in elliptic bounding boxes. Rather than learning an elliptic regressor, we directly regressed the rectangular boxes for faces, and compared our method with other methods \cite{bmf1, hr, bmf2, bmf3, bmf4,bmf5,bmf6,bmf7,bmf8,bmf9,dpm1, bmf10, bmf11, bmf12,bmf13,bmf14,bmf15,bmf16,bmf17,bmf18} which use no additional self-annotations. Our detector also achieves the top AP score of 97.9\% (depicted in Figure 4(c)).

\subsection{Runtime analysis}
\begin{figure}[t]
    \centering
    \includegraphics[width=0.32\textwidth]{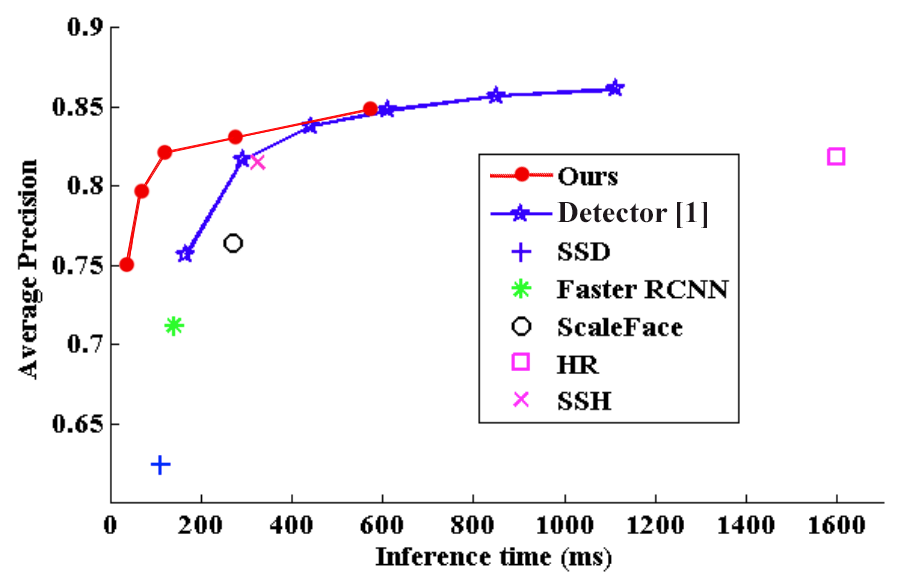}
    \caption{Inference times versus AP scores of our method and various methods on Wider Face Validation Hard set.}
\end{figure}
\setlength{\textfloatsep}{0.4cm}
Finally, we evaluated the inference time of our model. Following \cite{ssf}, we recorded the AP scores and the corresponding inference times of variously sized images with the same hardware configuration (a single NVIDIA Titan X GPU with a batch size of 1). Figure 5 shows that our model achieved very fast inference time with competitive accuracy compared with the detector in \cite{ssf}. In fact, the AP-versus-runtime curve of our model enveloped those of the others except that of \cite{ssf}. On the Wider Face Validation hard set, our detector forwarded one image within 36 ms and achieved 75.0\% AP score. Meanwhile, the detector in \cite{ssf} needs more than 150ms to process an image to obtain 75.7\% AP score. Our speed advantage is mainly gained from the single-stage design with one predefined anchor per cell whereas the detector in \cite{ssf} adopts a two-stage design with several predefined anchors per cell.
\section{CONCLUSIONS}
This work introduced a simple single-stage model for face detection which reasonably handles the speed-accuracy trade-off. Understanding the causes of the high false positive rate in single-stage methods was important for designing an effective, accuracy face detection framework. With SSD fashion, our model uses ResNet-101 as the feature extractor and incorporates several effective strategies that lower the rate of false positive together with fast inference time. Our model consistently achieved competitive results on common benchmarks for face detection with superiorly real-time inference.
\baselineskip 1.0\normalbaselineskip

\newpage
\section{Supplementary Materials}
\setlength{\textfloatsep}{1cm}
\subsection{The details of our baseline}
To avoid cluttering, we describe some details of our baseline here. The feature extractor of our baseline has the same architecture like that of our proposed model with the convolutional layers (ResNet-101), and extra convolutional layers. The detection layers extracted from the feature extractor are convolved directly with classification and regression conv layers. The anchor assigning and training on the model are followed that of the single shot detector (SSD).

\subsection{Precision-recall curves on Wider Face dataset}
In this section, we present the precision-recall curves on Wider Face validation and test set which are omitted in our paper. The results were obtained using the standard evaluation protocol (validation set) and from the test server (test set). The precision-recall curves on Wider Face validation set is shown in Figure 6 while the precision-recall curves on Wider Face test set is shown in Figure 7.

\subsection{Some qualitative results}
In this section, we show some qualitative results on several benchmark datasets (see in the next pages) in Figure VI.6 and Figure VI.7. We denote the red bounding boxes as the detected faces by our models. Best view the results in color.
\includegraphics[width=0.98\textwidth]{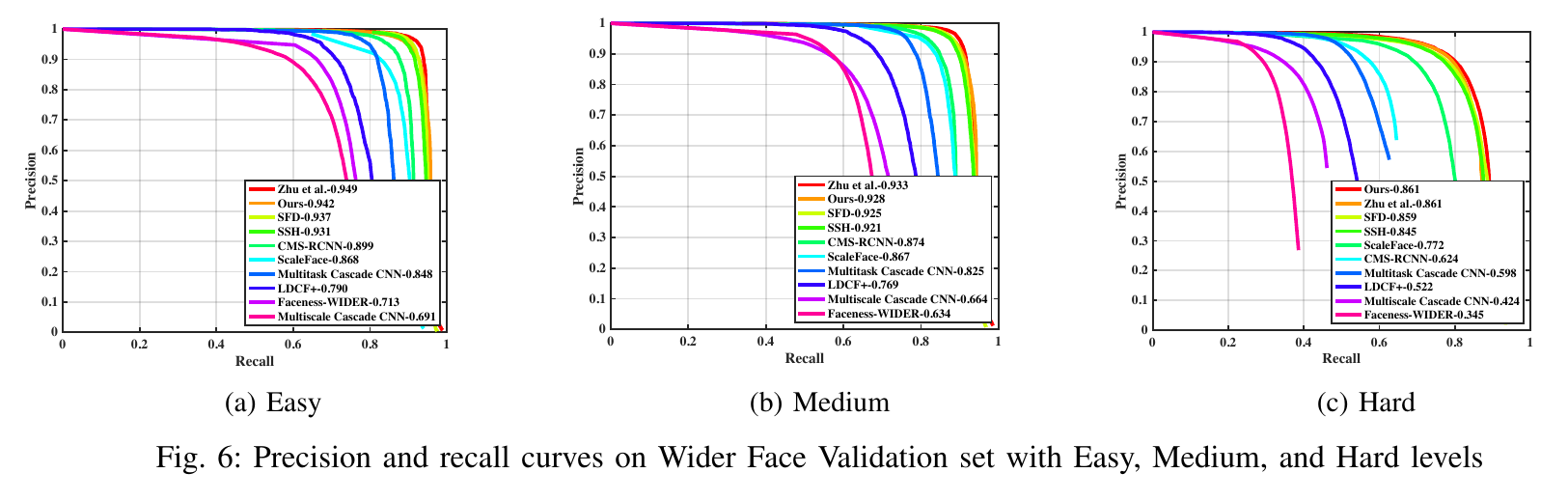}
\includegraphics[width=0.98\textwidth]{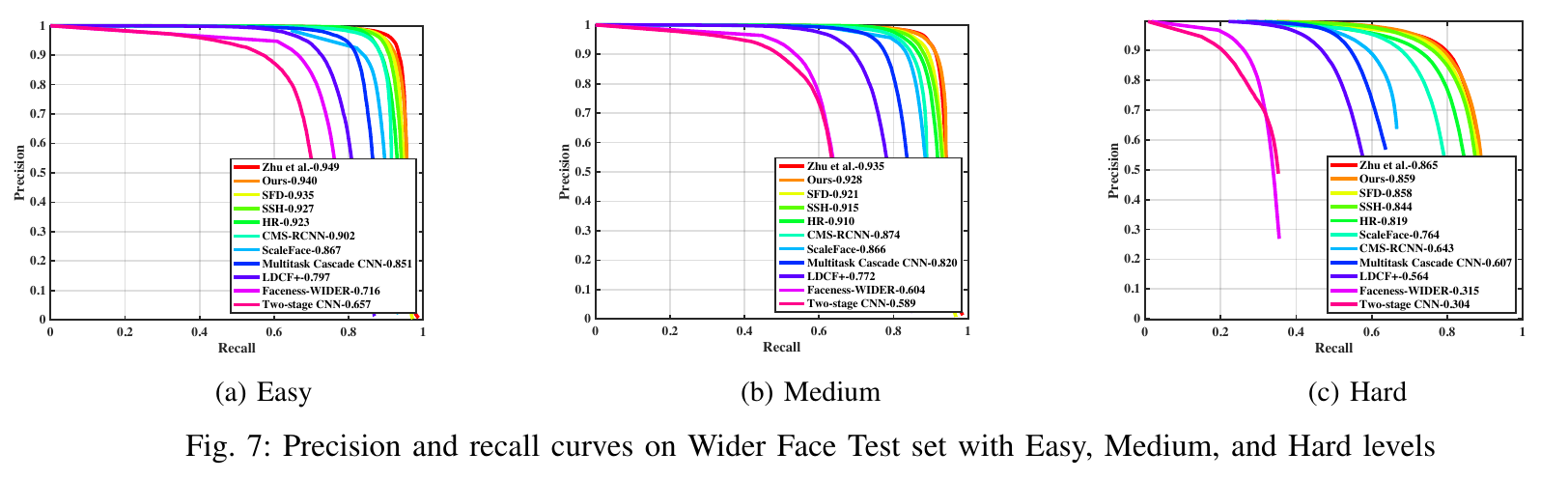}

\renewcommand\thefigure{\thesection.\arabic{figure}}  
\begin{figure*}[t]
     \centering
    \begin{subfigure}[t]{1.0\textwidth}
    \centering
    \includegraphics[width=0.45\textwidth]{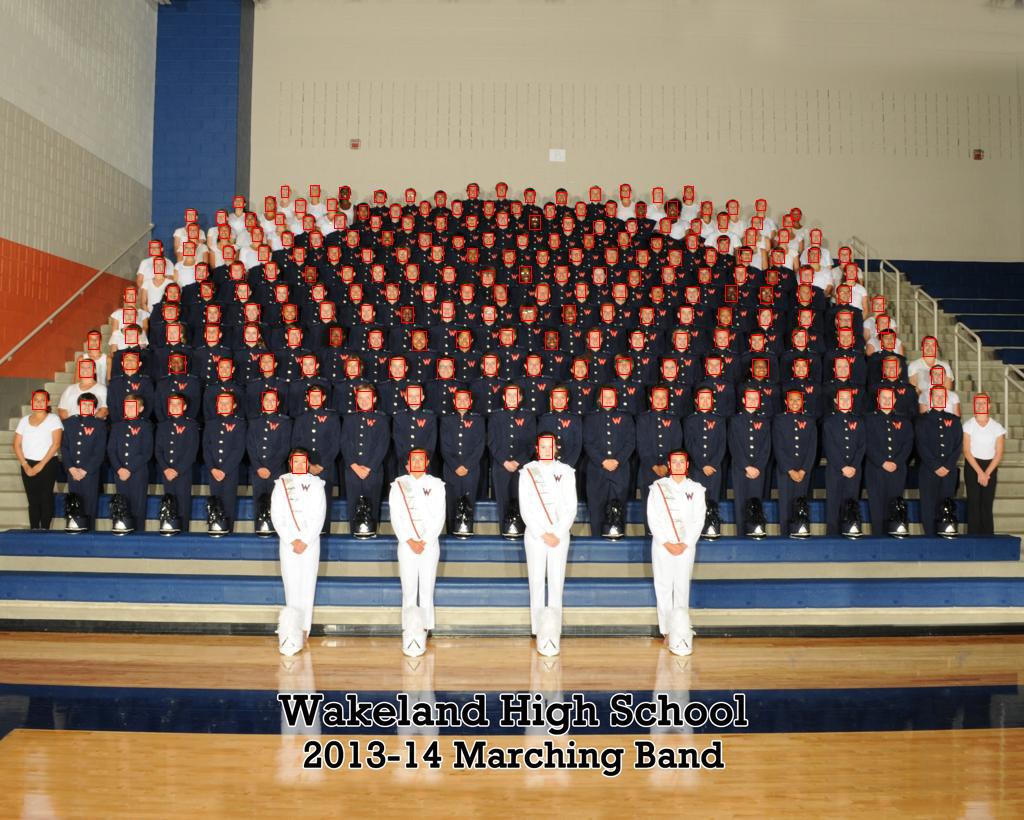}
    \includegraphics[width=0.45\textwidth]{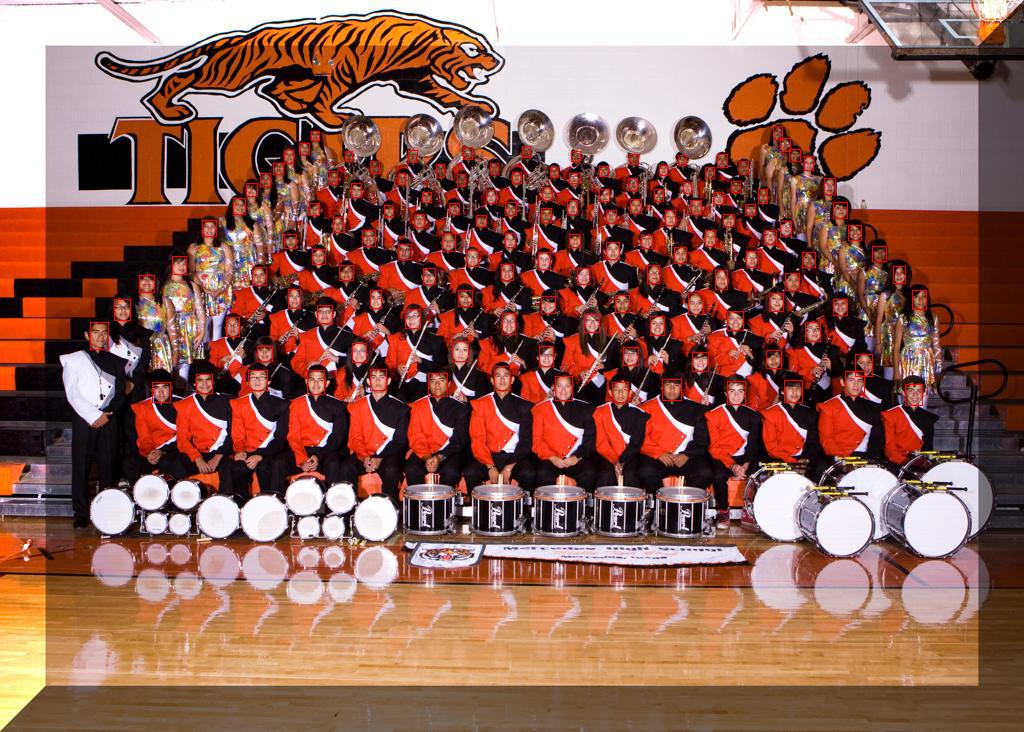}
    \includegraphics[width=0.45\textwidth]{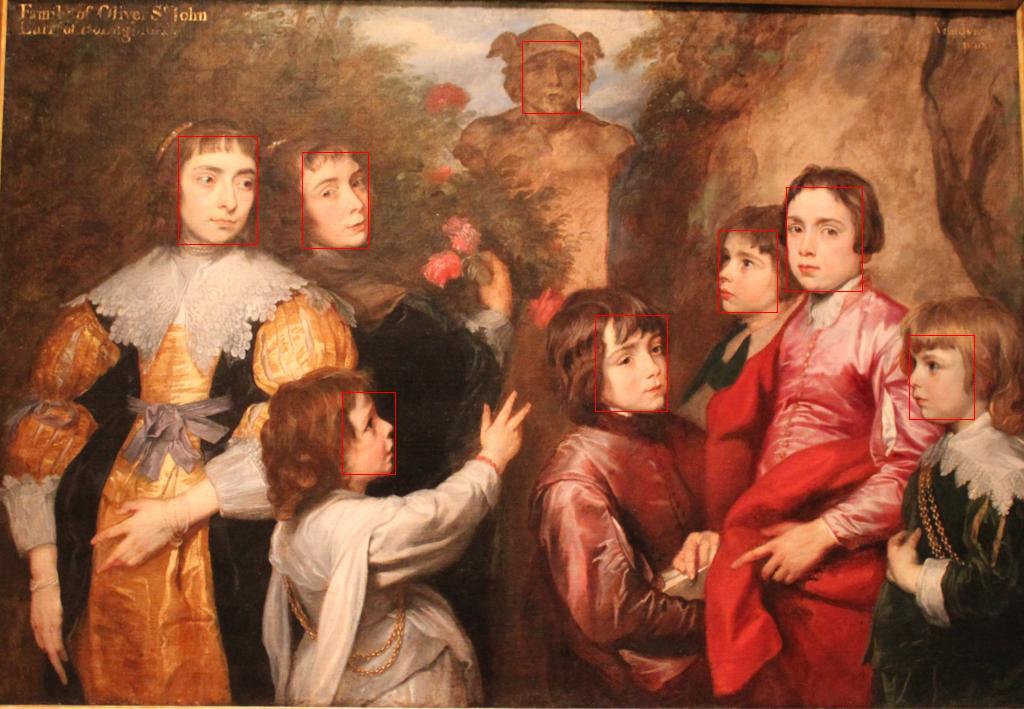}
    \includegraphics[width=0.45\textwidth]{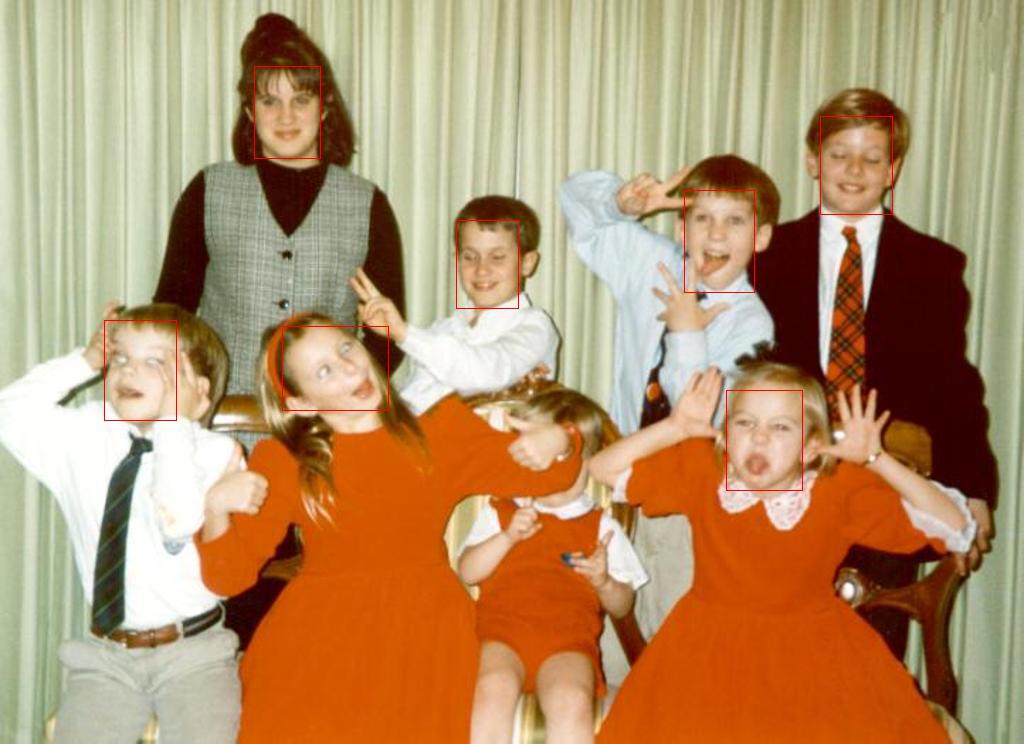}
    \includegraphics[width=0.45\textwidth]{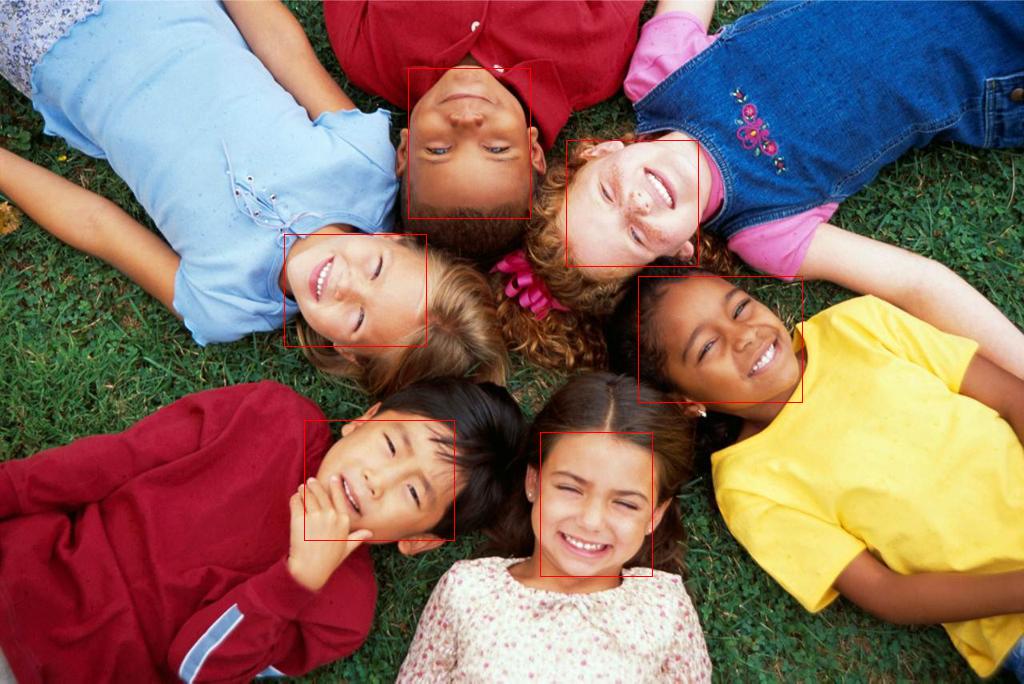}
    \includegraphics[width=0.45\textwidth]{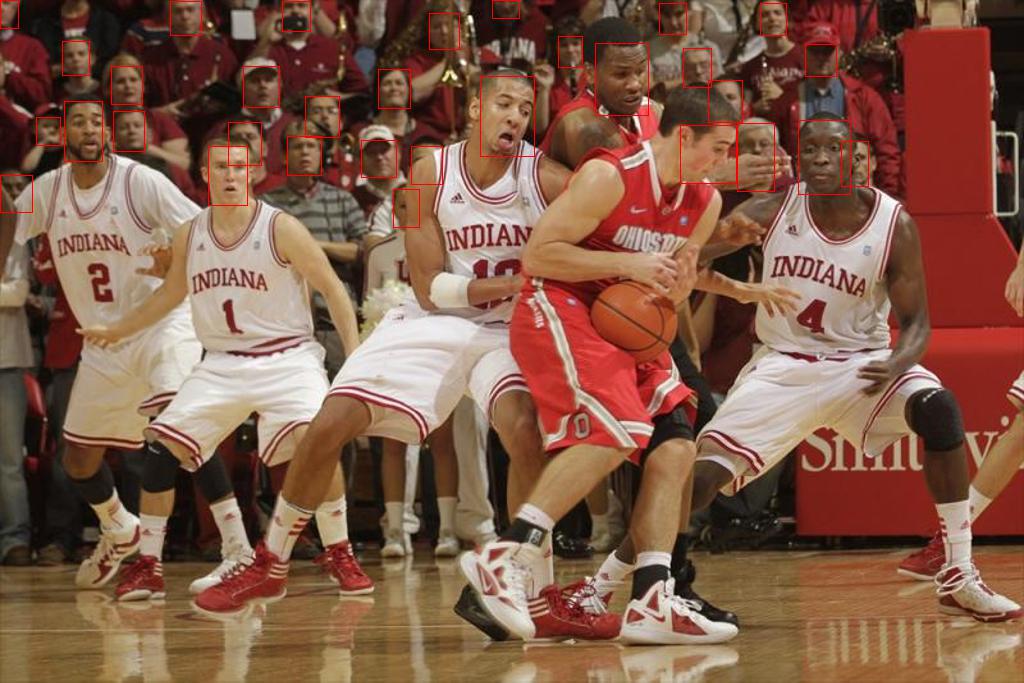}
    \includegraphics[width=0.45\textwidth]{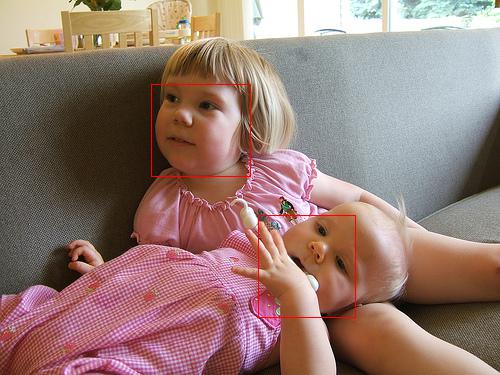}
    \includegraphics[width=0.45\textwidth]{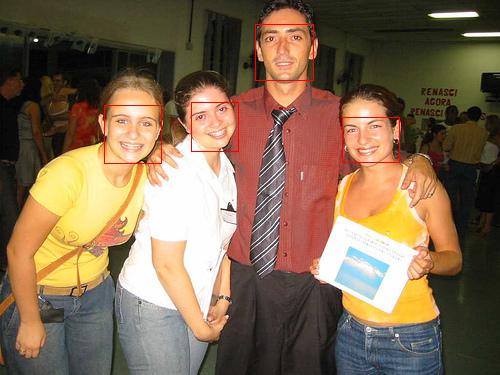}
    
    \end{subfigure}
    \caption{Several qualitative results on several datasets. Best view in color.}
\end{figure*}

\begin{figure*}[t]
     \centering
    \begin{subfigure}[t]{1.0\textwidth}
    \centering
    \includegraphics[width=0.45\textwidth]{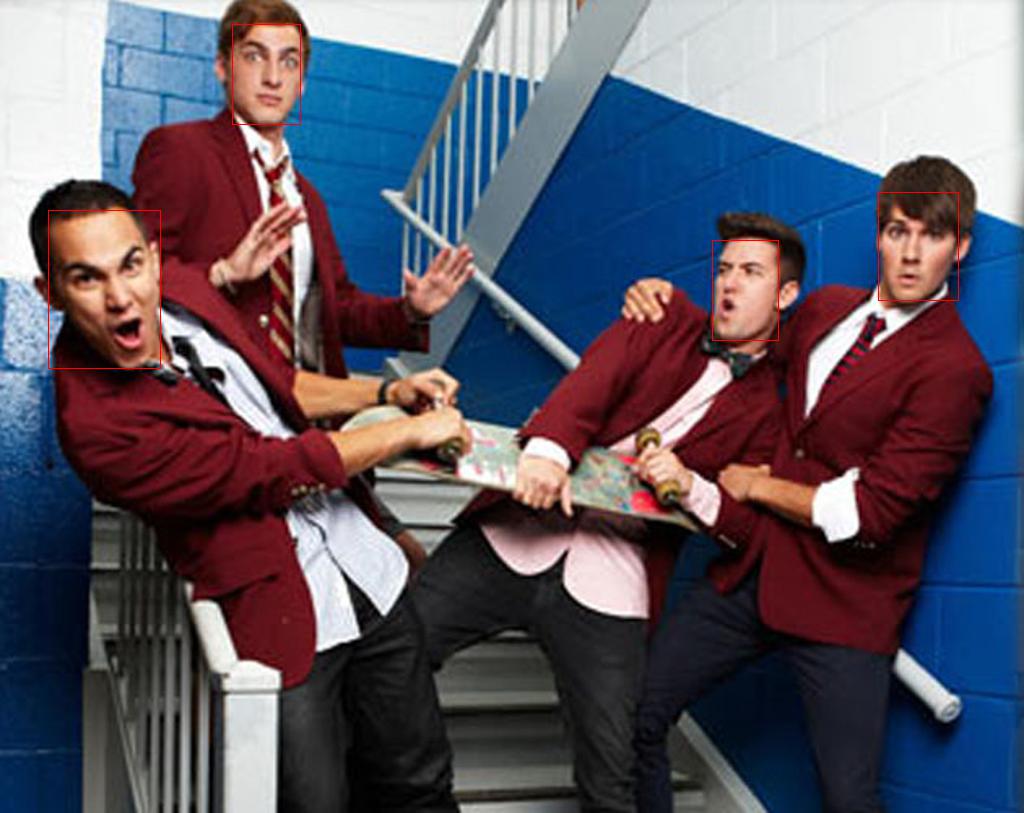}
    \includegraphics[width=0.45\textwidth]{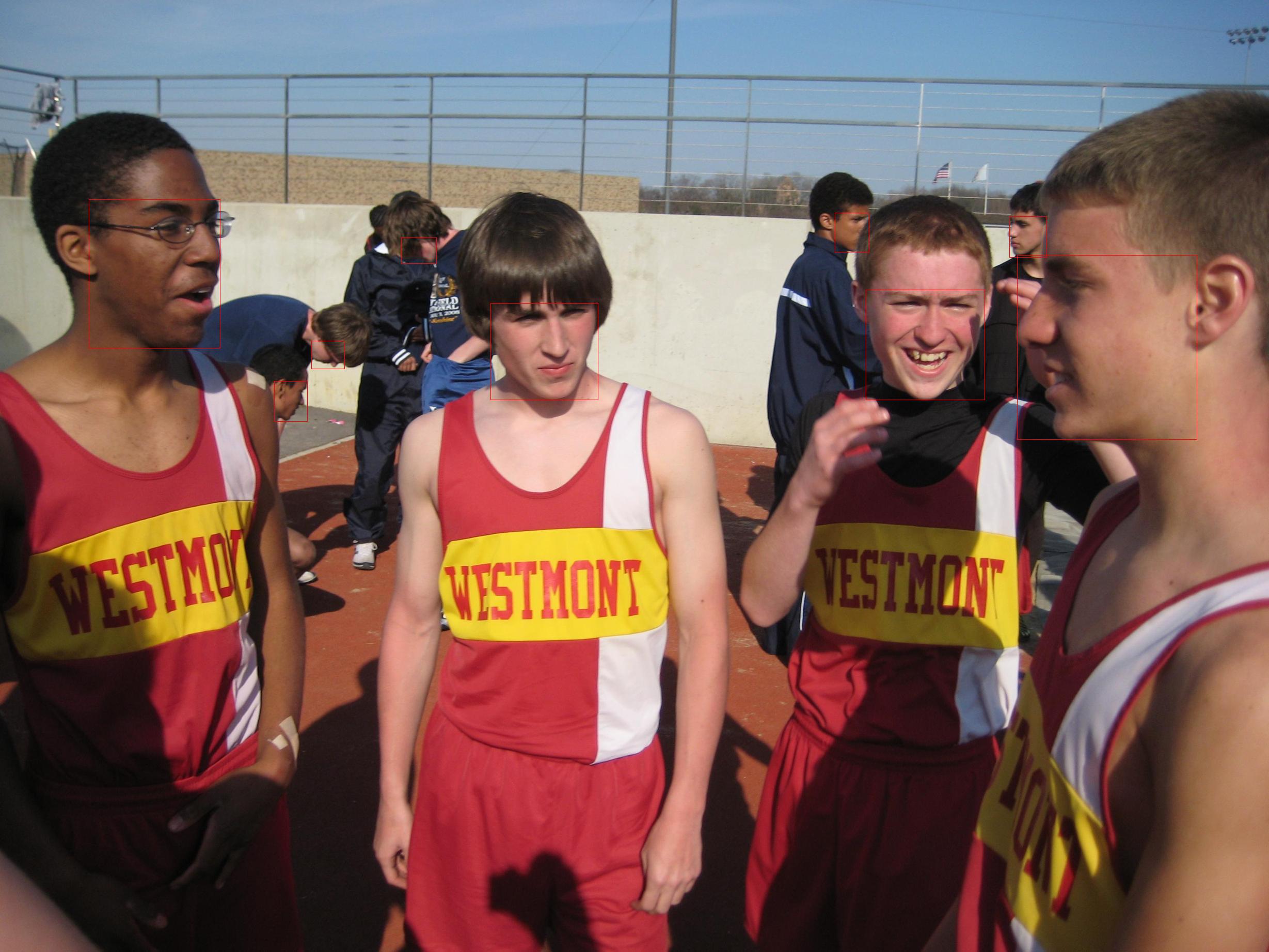}
    \includegraphics[width=0.45\textwidth]{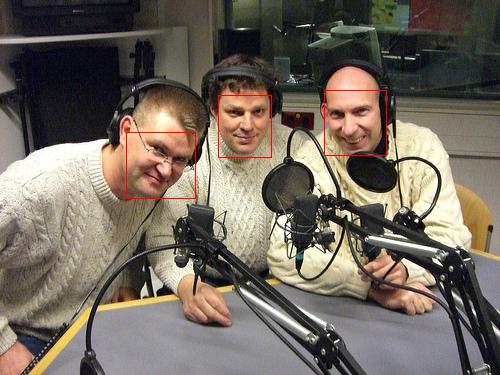}
    \includegraphics[width=0.45\textwidth]{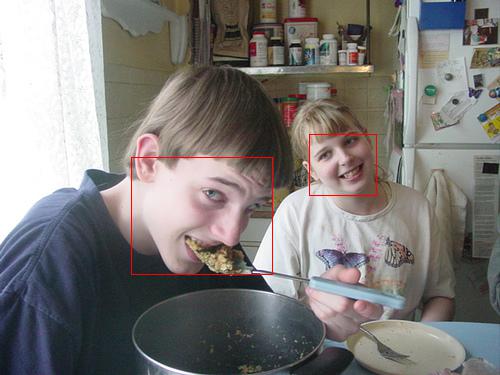}
    \includegraphics[width=0.45\textwidth]{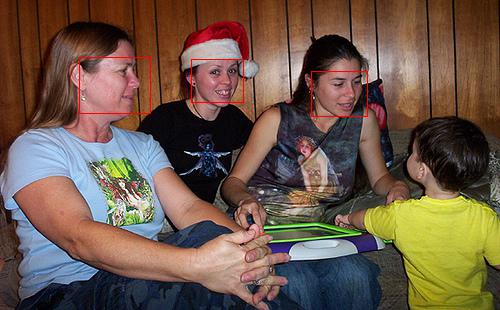}
    \includegraphics[width=0.45\textwidth]{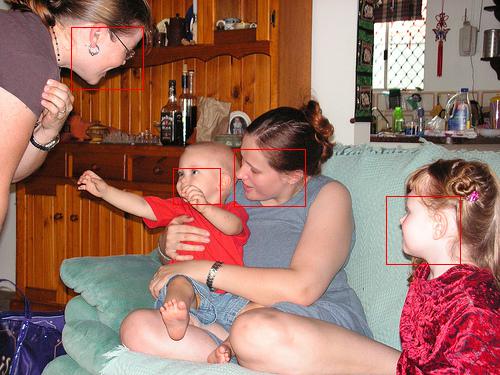}
    \includegraphics[width=0.45\textwidth]{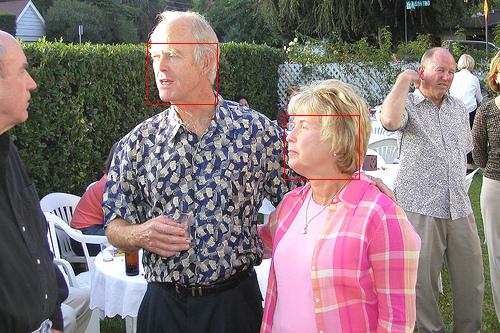}
    \includegraphics[width=0.45\textwidth]{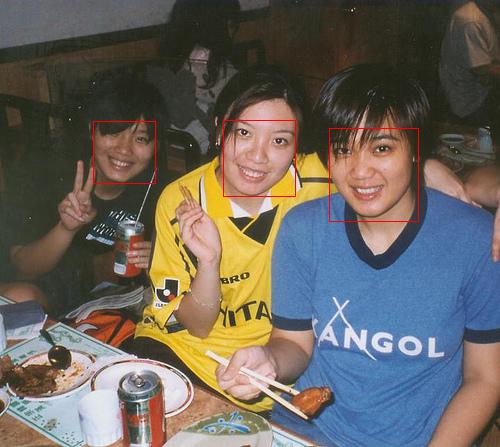}
    
    \end{subfigure}
    \caption{Several qualitative results on several datasets. Best view in color.}
\end{figure*}

\end{document}